%% file: iclr2027_conference.tex
\documentclass{article}
\usepackage{iclr2027_conference,times}
\input{math_commands.tex}

\usepackage{amsmath}
\usepackage{amssymb}
\usepackage{graphicx}
\usepackage{booktabs}
\usepackage{multirow}
\usepackage{algorithm}
\usepackage{algpseudocode}
\usepackage{float}
\usepackage{capt-of}
\usepackage{placeins}
\usepackage{microtype}
\usepackage{url}
\usepackage{hyperref}

\DeclareMathOperator*{\argmaxop}{arg\,max}
\newcommand{\method}{ChunkVAE}
\newcommand{\bbop}{BBOP}
\newcommand{\scurve}{S-Curve}

\title{Beyond Global Latents: Chunk-Based Sparse Grid VAE for Scalable 3D Modeling}

\author{
Kaiyi Zhang$^{1,2}$, Zhihao Liang$^{2}$, Haolin Liu$^{2}$, Qingxiang Lin$^{2}$, Zeqiang Lai$^{3,2}$, Yunfei Zhao$^{2}$, \\
\textbf{Bowen Zhang$^{2}$, Xianghui Yang$^{2}$, Zibo Zhao$^{2}$, Chunchao Guo$^{2\dagger}$, Long Quan$^{1\dagger}$} \\
\\
$^1$Hong Kong University of Science and Technology \\
$^2$Tencent Hunyuan \\
$^3$The Chinese University of Hong Kong \\
$^\dagger$Corresponding authors \\
\texttt{\{kzhangca, quan\}@cse.ust.hk}, \texttt{chunchaoguo@tencent.com}
}

\iclrfinalcopy
\begin{document}
\maketitle

\begin{abstract}
Sparse voxel grids preserve the spatial structure needed for detailed 3D reconstruction, but their memory still grows rapidly with resolution as active surface cells increase. We introduce \textit{\method}, a sparse grid variational autoencoder organized around local chunks rather than a global latent volume. Local learned operators permit independently chosen encoder and decoder partitions and allow inference chunk sizes to differ from training. Two complementary data operators make this flexibility practical: \textit{Balanced Binary Object Partitioning} distributes active cells while limiting replicated overlap, while \textit{\scurve{} weighted stitching} attenuates unreliable boundary features when assembling a global latent or reconstruction. Across three object benchmarks, \textit{\method} is competitive with or better than strong baselines from $512^3$ to $1536^3$; smaller chunks lower peak allocated memory and shorten per-chunk compute, enabling faster parallel inference. Stable stitched latents and improved image to 3D metrics indicate that local compression can scale geometry while retaining the global interface required downstream.
\end{abstract}

\section{Introduction}
\label{sec:intro}

High fidelity 3D generation relies on an autoencoder that preserves geometry before a generative model learns its latent distribution~\citep{zhang20233dshape2vecset,zhao2024michelangelo,li2024craftsman,zhang2024clay,hunyuan3d2025hunyuan3d,lai2025lattice,xiang2024structured,xiang2025native}. Errors introduced by this compression stage cannot be recovered downstream. Improving the spatial resolution and capacity of 3D variational autoencoders is therefore a direct route to sharper generated geometry.

Existing 3D VAEs commonly use global vector sets or sparse voxel grids. Vector set models compress a surface into global tokens and decode a continuous field~\citep{zhang20233dshape2vecset,chen2025dora,li2025triposg}. Their compact latent is attractive for generation, but local detail must be conveyed through a limited global token set. Sparse grid models retain explicit spatial support and have produced strong high resolution reconstructions~\citep{ren2024xcube,xiang2024structured,he2025sparseflex,wu2025direct3d,xiang2025native}. Sparsity removes empty space, but it does not remove the resolution dependence of active surface cells. Increasing the grid resolution consequently raises activation memory and restricts both model size and usable spatial extent.

\begin{figure}[t]
    \centering
    \includegraphics[width=\textwidth]{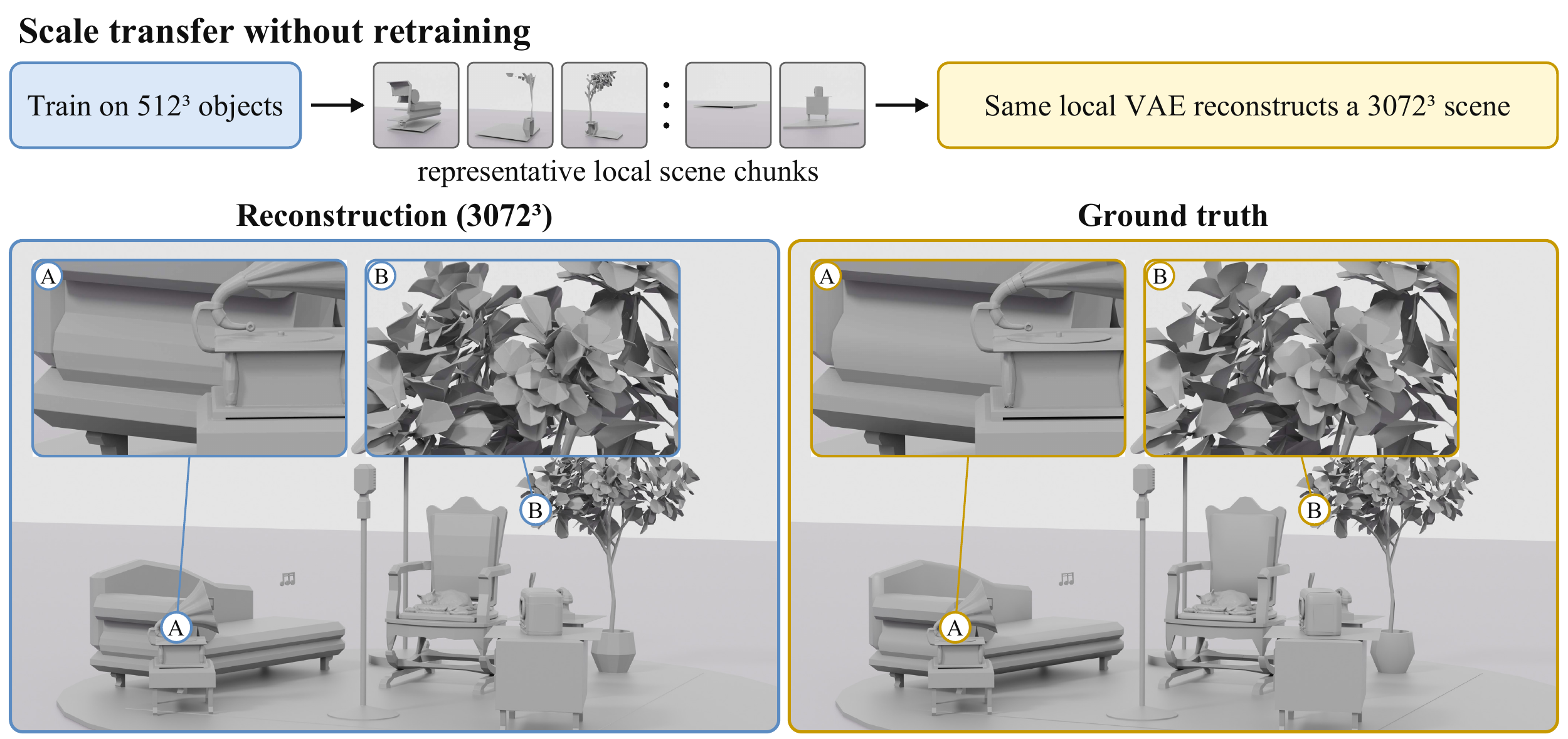}
    \caption{Resolution-flexible inference. Trained on chunks from $512^3$ objects, the same local VAE reconstructs a $3072^3$ scene without retraining. \textbf{\textit{Best viewed with zoom-in.}}}
    \label{fig:scene_infer}
\end{figure}

Chunking appears to offer a simple solution, yet it changes the representation learned by the VAE. Uniform spatial partitions can contain very different numbers of active cells, causing load imbalance and occasionally leaving a chunk too sparse to survive decoder pruning. Padding restores context but replicates data, and direct averaging transfers unstable edge features into visible seams. More importantly, methods that rely on global attention cannot freely change partition boundaries because a chunk no longer receives the same global context. Prior sparse grid systems therefore treat a global, unchunked encoder as necessary for preserving latent integrity, applying chunking primarily in the decoder or tying it to a fixed partition~\citep{ren2024xcube,he2025sparseflex,wu2025direct3d}.

Our starting point is a representation principle: a spatial latent can be assembled consistently from independently processed regions when both the learned operators and the assembly rule are local. Based on this principle, we develop \method, a sparse grid VAE that uses sparse convolutions and windowed attention throughout. Its encoder and decoder partitions are independent, and the inference chunk budget need not match the training budget. This separates three quantities that are otherwise coupled: global resolution, per-device memory, and training chunk size. Figure~\ref{fig:scene_infer} illustrates the resulting resolution flexibility at inference.

Two deterministic operators support this learned representation. Balanced Binary Object Partitioning (\bbop) recursively splits active cells at the median of their longest axis. It improves load balance without the large overlap induced by clustering. \scurve{} weighted stitching assigns low confidence to the exterior of padded regions and smoothly increases confidence toward each chunk core. The same operator assembles encoder features and decoder outputs.

Our contributions are:
\begin{itemize}
    \item We formulate chunk-wise sparse grid compression around local learned operators, enabling independent encoder and decoder partitions and different chunk budgets at training and inference.
    \item We introduce \bbop{} and \scurve{} stitching to address load imbalance, overlap, and boundary reliability. Their effects are measured using balance, replication, and reconstruction.
    \item We demonstrate state-of-the-art overall object reconstruction from $512^3$ to $1536^3$, alongside leading image to 3D generation metrics at $1536^3$, and quantify memory, runtime, and resolution trade-offs.
\end{itemize}

\section{Related Work}
\label{sec:related}

\paragraph{Global 3D latents.}
VecSet VAEs encode point samples into a fixed collection of latent vectors and decode occupancy or signed distance values at query points~\citep{zhang20233dshape2vecset,zhao2024michelangelo,li2024craftsman,zhang2024clay,li2025triposg,chen2025dora}. Scaling tokens, surface sampling, and query resolution improves detail, while the encoder and decoder still bridge discrete samples and a continuous field. VoxSet methods inject spatial queries to retain more locality~\citep{lai2025lattice,jia2025ultrashape}. These representations offer a compact global interface for diffusion or flow models, but increasing local detail raises the number of global tokens.

\paragraph{Sparse grid autoencoders.}
Dense voxel VAEs are limited by volumetric memory~\citep{cheng2023sdfusion,li2023diffusion}. XCube~\citep{ren2024xcube} and TRELLIS~\citep{xiang2024structured} use sparse grids to concentrate computation near surfaces. SparseFlex~\citep{he2025sparseflex}, Direct3D-S2~\citep{wu2025direct3d}, Sparc3D~\citep{li2025sparc3d}, and TRELLIS.2~\citep{xiang2025native} improve decoding, geometric representations, or high resolution generation. XCube, SparseFlex, and Direct3D-S2 process spatial chunks, but use uniform partitions and retain a global encoder. In contrast, our question is whether the latent itself can be encoded, assembled, repartitioned, and decoded locally without losing its downstream utility.

\paragraph{Latent 3D generation.}
Modern 3D generation moves expensive denoising or flow matching into autoencoder latents~\citep{peebles2023scalable,ma2024sit,batifol2025}. The latent may be a triplane~\citep{gupta20233dgen,lan2024ln3diff}, vector set~\citep{li2025step1x,hunyuan3d2025hunyuan3d}, or sparse grid~\citep{xiang2024structured,wu2025unilat3d}. \method{} preserves a stitched global latent for compatibility with these models while making the VAE computation local. The downstream generative model remains global in our current system.

\begin{figure}[t]
    \centering
    \includegraphics[width=\textwidth]{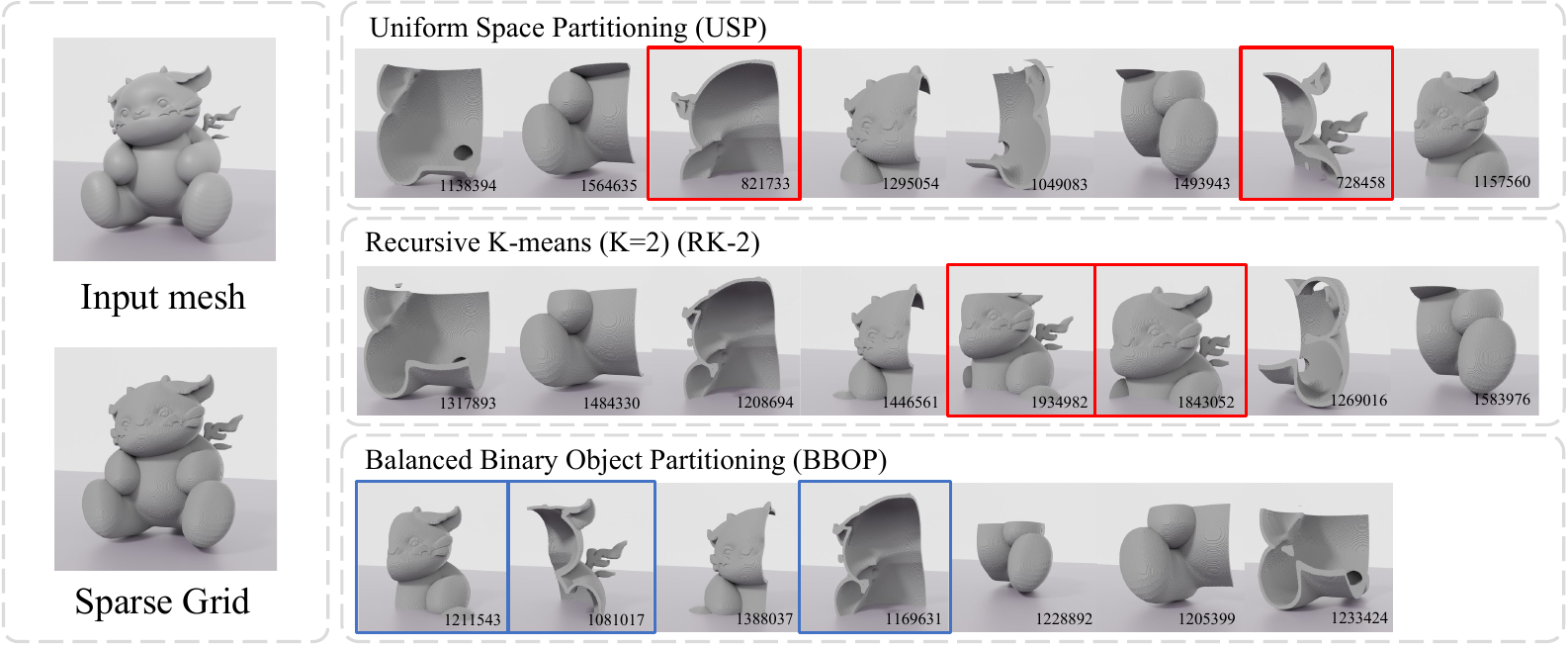}
    \caption{Partition strategies. Uniform Space Partitioning (USP) ignores the active cell distribution. Recursive K-means with $K=2$ (RK-2) balances counts but produces large overlapping boxes. \bbop{} balances active cells while retaining compact spatial support. Numbers indicate chunk sizes.}
    \label{fig:chunking}
\end{figure}

\section{Method}
\label{sec:method}

\subsection{Local Compression of a Global Sparse Grid}

Let a sparse grid at resolution $R$ be
\begin{equation}
    \mathcal{X}=\{(\mathbf{f}_j,\mathbf{c}_j)\}_{j=1}^{L},
    \qquad \mathbf{f}_j\in\mathbb{R}^{C},\quad
    \mathbf{c}_j\in\{0,\ldots,R-1\}^{3},
\end{equation}
where $\mathbf{f}_j$ is a scaled signed distance value and $\mathbf{c}_j$ is its coordinate. A partition operator produces overlapping chunks $\mathcal{P}(\mathcal{X})=\{\mathcal{X}_i\}_{i=1}^{N_c}$ whose active cell count is bounded by a budget $B$. An encoder $E$, a decoder $D$, and a stitching operator $\mathcal{S}$ yield
\begin{equation}
    \mathbf{Z}=\mathcal{S}\bigl(\{E(\mathcal{X}_i)\}_{i=1}^{N_c}\bigr),\qquad
    \hat{\mathcal{X}}=\mathcal{S}\bigl(\{D(\tilde{\mathbf{Z}}_k)\}_{k=1}^{N_d}\bigr),
    \label{eq:local_pipeline}
\end{equation}
where $\{\tilde{\mathbf{Z}}_k\}$ is a new partition of the stitched latent. Equation~\ref{eq:local_pipeline} permits $N_c\neq N_d$ and different boundaries only if $E$ and $D$ have bounded receptive fields and $\mathcal{S}$ handles boundary uncertainty. This requirement motivates both the network and the data operators below.

\subsection{Balanced Partitioning and Weighted Stitching}
\label{sec:partition}

\paragraph{Balanced Binary Object Partitioning.}
Figure~\ref{fig:chunking} compares the three strategies. USP divides the bounding volume uniformly, so its memory balance depends on shape geometry. RK-2 instead clusters active cells, but spatial bounding boxes around the clusters overlap substantially. \bbop{} preserves spatial coherence and balances active cells. For a set of coordinates, it selects the longest bounding box axis, splits at the median active coordinate, and recurses until each leaf contains at most $B$ cells. Each leaf box is padded to provide network context. Median splitting bounds the imbalance before padding, while longest-axis splitting keeps boxes compact. Algorithm~\ref{alg:bbop} is provided in Appendix~\ref{app:algorithms}.

\paragraph{\scurve{} weighted stitching.}
Features near the exterior of a padded chunk have less context than features near its core. Uniform averaging treats them as equally reliable. Let $[s_p,e_p]$ be a padded interval and $[s_c,e_c]$ its core. We use the smooth transition
\begin{equation}
S(t;p)=
\begin{cases}
\frac{1}{2}(2t)^p,&0\leq t<\frac{1}{2},\\
1-\frac{1}{2}\bigl(2(1-t)\bigr)^p,&\frac{1}{2}\leq t\leq1,
\end{cases}
\end{equation}
and define the one dimensional reliability
\begin{equation}
w(c)=
\begin{cases}
W_{\min}+\Delta W S\left(\frac{c-s_p}{s_c-s_p};p\right),&s_p\leq c<s_c,\\
W_{\max},&s_c\leq c<e_c,\\
W_{\max}-\Delta W S\left(\frac{c-e_c}{e_p-e_c};p\right),&e_c\leq c\leq e_p.
\end{cases}
\label{eq:weight}
\end{equation}
We set $p=2$, $W_{\min}=0.01$, and $W_{\max}=0.99$. The 3D weight is $W(\mathbf{c})=\prod_{d\in\{x,y,z\}}w(c_d)$, and duplicate features are combined by a scatter weighted average. This mitigates seams without aligned partitions or learned blending. Algorithm~\ref{alg:stitching} appears in Appendix~\ref{app:algorithms}; Figure~\ref{fig:stitching} compares padded unique selection, full and restricted means, and \scurve{} weighting with $p\in\{1,2\}$.

\subsection{\method{} Architecture}
\label{sec:architecture}

\begin{figure}[t]
    \centering
    \includegraphics[width=\textwidth]{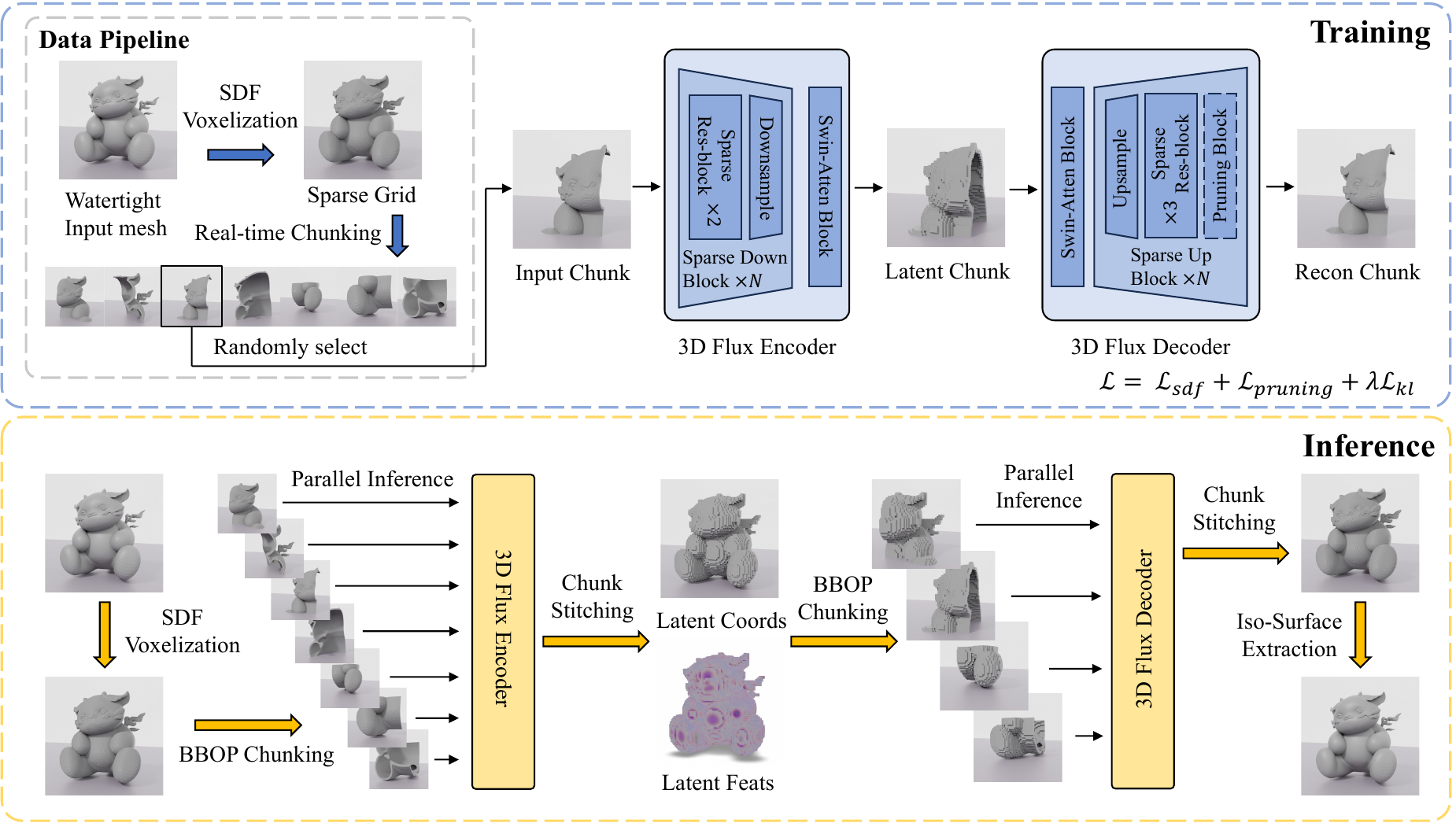}
    \caption{\method{} uses local operators during training and inference. Training reconstructs sampled chunks. At inference, encoder chunks are stitched into a global latent, repartitioned independently, decoded, and stitched into the output grid.}
    \label{fig:architecture}
\end{figure}

Figure~\ref{fig:architecture} summarizes training and inference. \method{} adapts the Flux image VAE~\citep{batifol2025} to the sparse framework of TRELLIS~\citep{xiang2024structured}, using three sparse-convolution downsampling and upsampling stages with supervised pruning. Replacing global bottleneck attention by one window-8 Swin block keeps every learned interaction local; this bounded receptive field, rather than the specific backbone, enables repartitioning.

Training samples one \bbop{} chunk and reconstructs the same spatial region. The objective is
\begin{equation}
\mathcal{L}=\mathcal{L}_{\mathrm{sdf}}+
\sum_{i=1}^{3}\mathcal{L}_{\mathrm{prune}}^{i}
+\lambda\mathcal{L}_{\mathrm{KL}},\qquad \lambda=10^{-6}.
\end{equation}
At inference, the encoder and decoder follow Equation~\ref{eq:local_pipeline}. Larger decoder chunks reduce stitched boundaries, while smaller chunks reduce peak memory. Section~\ref{sec:ablation} evaluates reconstruction and latent consistency under changed budgets and partitions.

\section{Experiments}
\label{sec:experiments}

\subsection{Setup and Evaluation Protocol}

\paragraph{Data and training.}
We convert meshes to watertight surfaces using flood filling~\citep{li2025sparc3d} followed by least-squares optimization of geometric details, then voxelize a narrow SDF band extending two cells on each side of the surface. Each resolution uses watertight voxelization at that same resolution. Training uses about 500K assets curated from Objaverse and Objaverse-XL~\citep{deitke2023objaverse,deitke2023objaverse_xl}. We first train at $512^3$, then fine-tune checkpoints at $1024^3$ and $1536^3$. The main model uses 32 GPUs, a total batch size of 64, AdamW with initial learning rate $10^{-4}$ and cosine decay, FP16 mixed precision, and 200K iterations. It reaches $8\times$ spatial compression in about 3.5 days. Architecture details appear in Appendix~\ref{app:implementation}.

\paragraph{Benchmarks and metrics.}
We evaluate on Toys4K~\citep{stojanov2021using}, ABO~\citep{collins2022abo}, and 95 complex High-quality Models (HQM) that do not overlap with training. Metrics such as Chamfer Distance (CD), Absolute Normal Consistency (ANC), and F-score at 0.001 are computed using point-to-surface distances following LATTICE~\citep{lai2025lattice}. All methods are evaluated against the same watertight target surfaces. VecSet and VoxSet baselines use 4096 latent tokens and $512^3$ marching cubes. Sparse grid inputs and extracted meshes use the reported evaluation resolution.

\subsection{Reconstruction and Resolution Scaling}

\begin{table}[h]
\caption{Reconstruction across resolutions. Metrics are CD ($\times10^4$), ANC ($\times10^2$), and F-score ($\times10^2$). At $512^3$, the 175M and 212M models are trained with chunk budgets below 8M and 1M cells, respectively. \textit{$^\dagger$ denotes our re-implementation.}}
\label{tab:reconstruction}
\centering
\scriptsize
\setlength{\tabcolsep}{1.8pt}
\renewcommand{\arraystretch}{0.92}
\resizebox{\textwidth}{!}{
\begin{tabular}{c l ccc ccc ccc}
\toprule
\multirow{2}{*}{Resolution} & \multirow{2}{*}{Method} & \multicolumn{3}{c}{Toys4K} & \multicolumn{3}{c}{ABO} & \multicolumn{3}{c}{HQM}\\
\cmidrule(lr){3-5}\cmidrule(lr){6-8}\cmidrule(lr){9-11}
&& CD$\downarrow$&ANC$\uparrow$&F$\uparrow$&CD$\downarrow$&ANC$\uparrow$&F$\uparrow$&CD$\downarrow$&ANC$\uparrow$&F$\uparrow$\\
\midrule
\multirow{8}{*}{$512^3$}
&Dora~\citep{chen2025dora} &7.52&99.07&94.69&6.54&99.56&95.61&13.47&96.91&82.36\\
&Hunyuan3D-2.1~\citep{hunyuan3d2025hunyuan3d} &8.02&98.96&92.08&6.01&99.45&96.22&14.20&96.49&79.38\\
&SparseFlex~\citep{he2025sparseflex} &2.49&99.46&99.16&2.34&99.70&99.59&4.48&98.30&96.82\\
&Direct3D-S2~\citep{wu2025direct3d} &5.83&98.98&97.22&4.74&99.47&98.74&9.60&96.79&88.90\\
&LATTICE$^\dagger$~\citep{lai2025lattice} &5.33&99.06&95.72&4.32&99.54&96.77&14.28&95.12&82.37\\
&TRELLIS.2~\citep{xiang2025native} &\underline{1.45}&99.45&99.61&\underline{1.04}&99.72&99.89&3.43&98.24&98.32\\
&\method{} (175M) &1.60&\underline{99.69}&\underline{99.64}&1.21&\underline{99.88}&\underline{99.94}&\underline{3.00}&\underline{98.89}&\underline{98.59}\\
&\method{} (212M) &\textbf{1.21}&\textbf{99.77}&\textbf{99.78}&\textbf{1.03}&\textbf{99.91}&\textbf{99.96}&\textbf{2.44}&\textbf{99.14}&\textbf{99.18}\\
\midrule
\multirow{4}{*}{$1024^3$}
&SparseFlex~\citep{he2025sparseflex} &1.28&99.64&99.68&1.28&99.80&99.88&2.38&98.59&98.75\\
&Direct3D-S2~\citep{wu2025direct3d} &3.38&99.37&99.47&3.19&99.63&99.76&4.32&97.83&97.79\\
&TRELLIS.2~\citep{xiang2025native} &\underline{0.47}&\underline{99.70}&\underline{99.98}&\textbf{0.35}&\underline{99.84}&\underline{99.99}&\underline{1.20}&\underline{98.81}&\textbf{99.89}\\
&\method{} (212M) &\textbf{0.45}&\textbf{99.89}&\textbf{99.98}&\underline{0.67}&\textbf{99.95}&\textbf{99.99}&\textbf{0.90}&\textbf{99.45}&\underline{99.88}\\
\midrule
\multirow{2}{*}{$1536^3$}
&TRELLIS.2~\citep{xiang2025native} &\underline{0.40}&\underline{99.78}&\underline{99.99}&\textbf{0.27}&\underline{99.86}&\underline{99.99}&\underline{0.67}&\underline{99.04}&\underline{99.97}\\
&\method{} (212M) &\textbf{0.36}&\textbf{99.92}&\textbf{99.99}&\underline{0.39}&\textbf{99.97}&\textbf{99.99}&\textbf{0.53}&\textbf{99.58}&\textbf{99.97}\\
\bottomrule
\end{tabular}}
\end{table}

\begin{figure}[H]
    \centering
    \includegraphics[width=\textwidth]{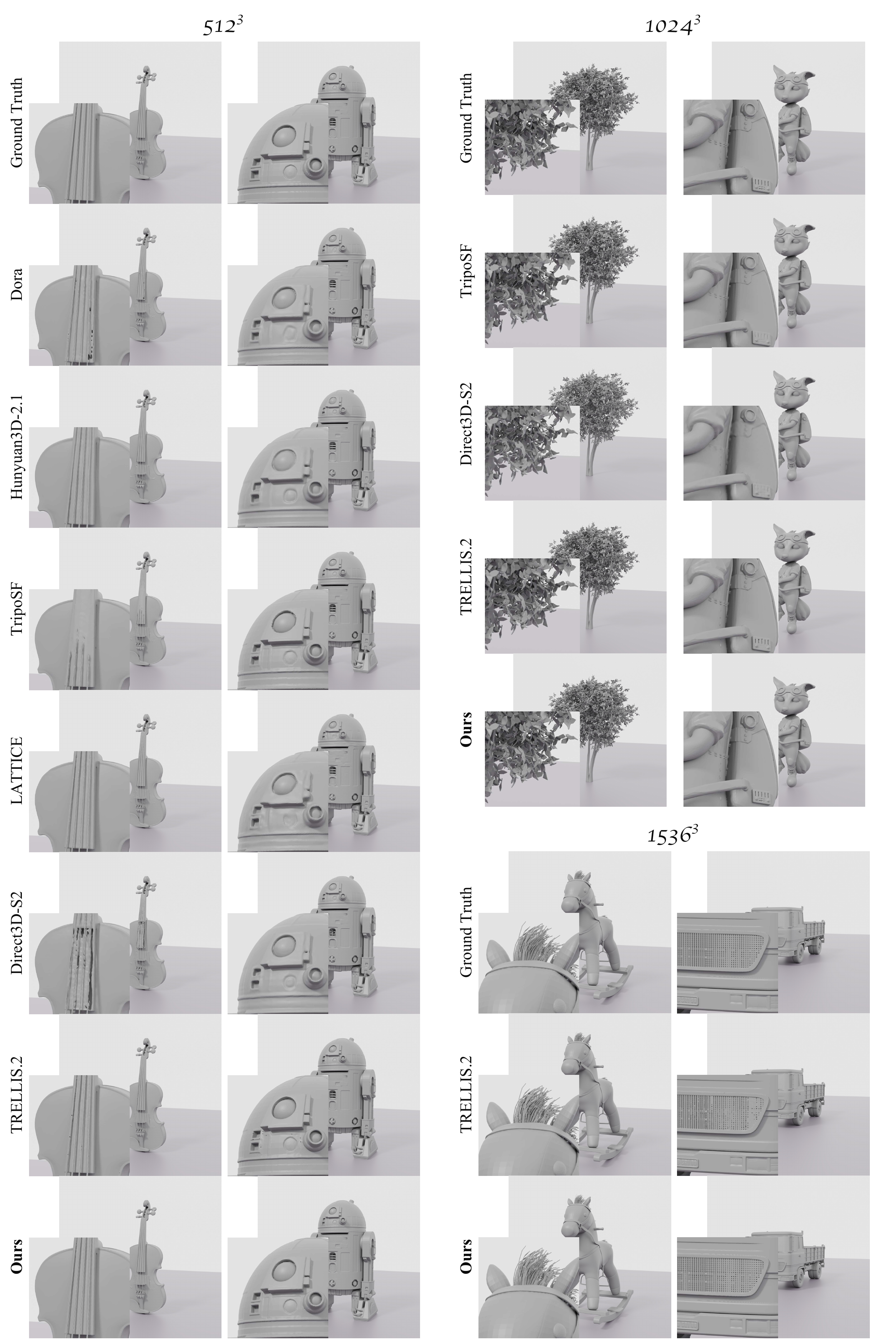}
    \caption{Qualitative Toys4K reconstruction at $512^3$, $1024^3$, and $1536^3$. \textit{Best viewed with zoom-in.}}
    \label{fig:reconstruction}
\end{figure}

Table~\ref{tab:reconstruction} shows that \method{} delivers state-of-the-art overall reconstruction quality across three benchmarks and resolutions from $512^3$ to $1536^3$. Each resolution is evaluated against separately generated, resolution-matched ground truth, and baseline results use the checkpoints released by the corresponding methods. As shown in Figure~\ref{fig:reconstruction}, \method{} preserves thin structures such as cello strings, fine front-grille lattices, horse manes, and tree crowns with sharper, more continuous geometry than the baselines, while avoiding the holes visible in TRELLIS.2 reconstructions.

\clearpage

\subsection{Efficiency}

\begin{table}[t]
\caption{Average $512^3$ inference on HQM objects using one GPU. Parentheses give mean time per chunk, an idealized parallel compute time when enough workers are available; scheduling and communication are excluded. Memory is peak allocated memory.}
\label{tab:efficiency}
\centering
\small
\setlength{\tabcolsep}{3pt}
\resizebox{\textwidth}{!}{
\begin{tabular}{l ccc cccc}
\toprule
\multirow{2}{*}{Metric}&\multirow{2}{*}{SparseFlex}&\multirow{2}{*}{Direct3D-S2}&\multirow{2}{*}{TRELLIS.2}&\multicolumn{4}{c}{\method{} cell budget}\\
\cmidrule(lr){5-8}
&&&&${<}1$M&${<}2$M&${<}4$M&No chunking\\
\midrule
Time (s)$\downarrow$&8.43&22.07&14.90&10.46 (\textbf{2.34})&7.61 (3.29)&6.19 (4.67)&\textbf{5.71}\\
Memory (MB)$\downarrow$&7267.60&21226.11&\textbf{2668.47}&\textbf{4569.84}&6394.03&10244.87&16219.47\\
\bottomrule
\end{tabular}}
\end{table}

\begin{figure}[t]
\centering
\captionof{table}{Image to 3D evaluation following the LATTICE protocol.}
\label{tab:generation}
\small
\setlength{\tabcolsep}{4pt}
\begin{tabular}{l cccc}
\toprule
Method&ULIP-T$\uparrow$&ULIP-I$\uparrow$&Uni3D-T$\uparrow$&Uni3D-I$\uparrow$\\
\midrule
CraftsMan~\citep{li2024craftsman}&0.074&0.129&0.237&0.298\\
Hunyuan3D-2.1~\citep{hunyuan3d2025hunyuan3d}&0.077&\underline{0.130}&0.251&0.315\\
Direct3D-S2~\citep{wu2025direct3d}&0.074&0.122&0.247&0.314\\
LATTICE~\citep{lai2025lattice}&\underline{0.078}&\underline{0.130}&\underline{0.254}&0.315\\
TRELLIS.2~\citep{xiang2025native}&0.077&0.124&0.245&\underline{0.317}\\
\method{} ($1536^3$)&\textbf{0.079}&\textbf{0.134}&\textbf{0.255}&\textbf{0.318}\\
\bottomrule
\end{tabular}
\vspace{5pt}
\includegraphics[width=\textwidth]{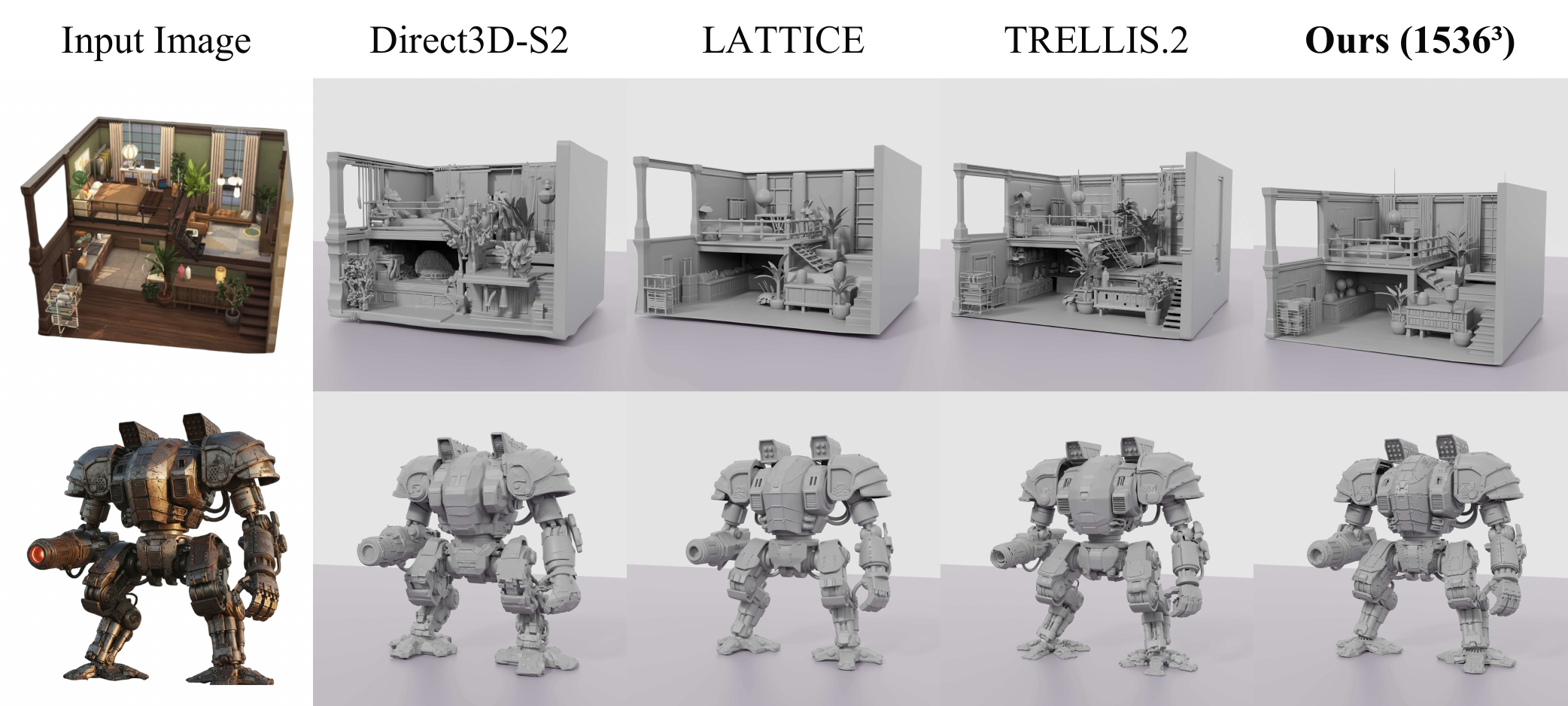}
\captionof{figure}{Image to 3D comparison with in-the-wild images. \textit{Best viewed with zoom-in.}}
\label{fig:generation}
\end{figure}

Table~\ref{tab:efficiency} compares inference time and peak allocated memory at $512^3$. \method{} is faster than Direct3D-S2 and TRELLIS.2 under every tested budget: it takes 5.71 s without chunking and 10.46 s even below 1M cells, versus 22.07 s and 14.90 s for the two baselines. The ${<}2$M setting also improves over SparseFlex in both latency (7.61 versus 8.43 s) and memory (6.39 versus 7.27 GB). Within \method{}, reducing the budget from no chunking to ${<}1$M lowers memory by 71.8\%, from 16.22 to 4.57 GB, while independent chunks reduce idealized parallel compute to 2.34 s per chunk. TRELLIS.2 uses fewer cells because its o-voxel represents only a surface layer; our SDF grid extends two cells inside and outside the surface to maintain watertight continuity and avoid the hole artifacts visible in the TRELLIS.2 reconstructions in Figure~\ref{fig:reconstruction}, naturally requiring more memory. Overall, \method{} provides the strongest measured balance of inference speed and adjustable memory, with ${<}2$M serving as a practical default.

\subsection{Image to 3D Generation}

We use the two stage flow matching pipeline of TRELLIS~\citep{xiang2024structured}, replacing stage two's projected DINOv2 features with \method{} SDF latents; sampling uses 25 steps and classifier-free guidance 7.5. Table~\ref{tab:generation} and Figure~\ref{fig:generation} provide quantitative and qualitative comparisons, respectively. Under the LATTICE protocol, \method{} achieves state-of-the-art scores, improving over the strongest baseline by $+0.004$ on ULIP-I and $+0.001$ on Uni3D-I. Qualitatively, our generated shapes contain richer and more complete details, including steps and bookshelves in the indoor scene and rivets and articulated joints on the robot, while aligning more faithfully with the input images.

\subsection{What Makes Chunking Consistent?}
\label{sec:ablation}

\paragraph{Partition quality.}
For chunk sizes $s_i$ with mean $\mu$, Chunk Size Coefficient of Variation is
$\mathrm{CSCV}=\mu^{-1}\sqrt{N_c^{-1}\sum_i(s_i-\mu)^2}$.
Chunk Replication Factor is
$\mathrm{CRF}=\sum_i|c_i|/|\cup_i c_i|$.
Lower CSCV indicates better load balance and CRF close to one indicates less overlap. Success Ratio is the fraction of test shapes producing a complete reconstruction. Table~\ref{tab:partition} shows that \bbop{} reconstructs every shape at every resolution on all three datasets. The distinction is largest on HQM at $1536^3$: USP succeeds on only 0.800 of the shapes, while RK-2 reaches 0.989 and \bbop{} reaches 1.000. At that resolution \bbop{} also reduces CSCV to 0.150, compared with 0.876 for USP, and avoids the high replication of RK-2 (CRF 1.702 versus 2.584). Thus, \bbop{} achieves the best trade-off between balanced chunk sizes and limited overlap while preserving complete reconstructions.

\begin{table}[t]
\caption{Partitioning on Toys4K, ABO, and HQM. Each resolution group reports CSCV, CRF, and Success Ratio. CRF is bounded below by one, so lower is better.}
\label{tab:partition}
\centering
\small
\setlength{\tabcolsep}{1.7pt}
\begin{tabular}{l@{\hspace{2.5pt}}l ccc ccc ccc}
\toprule
\multirow{2}{*}{Dataset}&\multirow{2}{*}{Method}&\multicolumn{3}{c}{$512^3$}&\multicolumn{3}{c}{$1024^3$}&\multicolumn{3}{c}{$1536^3$}\\
\cmidrule(lr){3-5}\cmidrule(lr){6-8}\cmidrule(lr){9-11}
&&CSCV$\downarrow$&CRF$\downarrow$&Succ.~Ratio$\uparrow$&CSCV$\downarrow$&CRF$\downarrow$&Succ.~Ratio$\uparrow$&CSCV$\downarrow$&CRF$\downarrow$&Succ.~Ratio$\uparrow$\\
\midrule
\multirow{3}{*}{Toys4K}
&USP&0.220&2.058&1.000&0.644&1.967&0.974&0.684&1.951&0.927\\
&RK-2&0.082&1.344&1.000&0.194&1.815&1.000&0.204&1.954&0.998\\
&\bbop{}&\textbf{0.031}&\textbf{1.191}&\textbf{1.000}&\textbf{0.103}&\textbf{1.426}&\textbf{1.000}&\textbf{0.127}&\textbf{1.499}&\textbf{1.000}\\
\midrule
\multirow{3}{*}{ABO}
&USP&0.260&1.648&1.000&0.642&1.865&0.990&0.603&1.904&0.949\\
&RK-2&0.112&1.364&1.000&0.214&1.725&1.000&0.223&1.802&0.990\\
&\bbop{}&\textbf{0.053}&\textbf{1.276}&\textbf{1.000}&\textbf{0.137}&\textbf{1.510}&\textbf{1.000}&\textbf{0.156}&\textbf{1.558}&\textbf{1.000}\\
\midrule
\multirow{3}{*}{HQM}
&USP&0.318&1.950&1.000&0.884&1.887&0.989&0.876&2.303&0.800\\
&RK-2&0.137&1.797&1.000&0.230&2.437&1.000&0.226&2.584&0.989\\
&\bbop{}&\textbf{0.052}&\textbf{1.315}&\textbf{1.000}&\textbf{0.137}&\textbf{1.625}&\textbf{1.000}&\textbf{0.150}&\textbf{1.702}&\textbf{1.000}\\
\bottomrule
\end{tabular}
\end{table}

\begin{figure}[t]
    \centering
    \includegraphics[width=\textwidth]{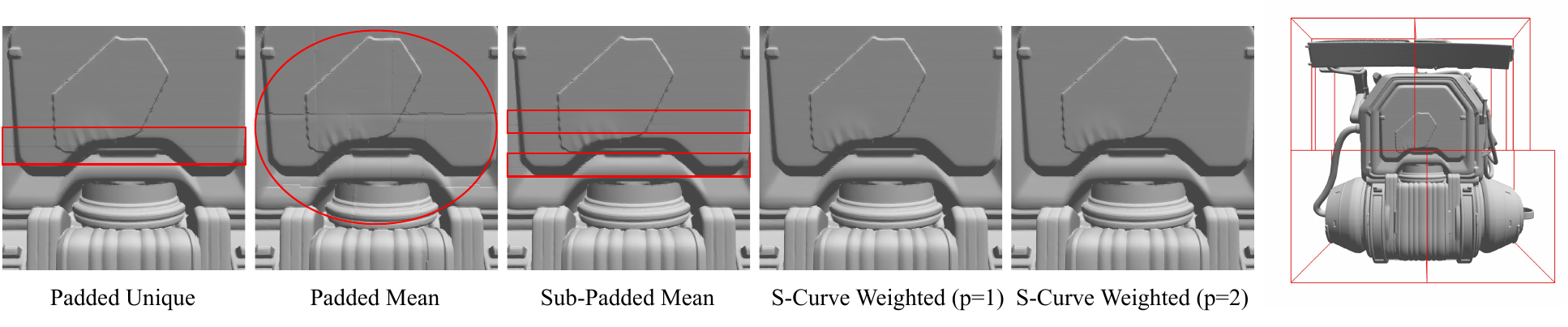}
    \caption{Stitching ablation under a fixed partition and trained VAE. Red boxes identify boundary artifacts. \scurve{} weighting reduces the contribution of features near padded exteriors.}
    \label{fig:stitching}
\end{figure}

\paragraph{Boundary reliability.}
Partition quality and stitching answer different questions: Table~\ref{tab:partition} measures workload and overlap, whereas Figure~\ref{fig:stitching} fixes the partition and VAE and changes only the merge rule. Padded unique selection, full averaging, and restricted averaging retain boundary artifacts. Uniform averaging gives boundary and interior predictions equal weight, allowing unreliable boundary features to contaminate the fused result. Restricting the averaging region removes these features but introduces a hard cutoff where the set of contributing chunks changes, potentially creating new seams. \scurve{} weighting instead retains the overlap and varies each contribution smoothly with available context; $p=2$ suppresses exterior features more strongly than the displayed $p=1$ variant.

\paragraph{Chunk and partition decoupling.}
Table~\ref{tab:decoupling} evaluates decoupling across training budget, inference budget, and encoder--decoder partition choice. In the left panel, changing the training budget from ${<}1$M to ${<}100$K at any fixed inference budget gives $|\Delta\mathrm{CD}|\leq0.08$, showing that inference chunks need not match those used for training. Increasing the inference budget from ${<}1$M to ${<}8$M improves CD from 2.99/3.04 to 2.42/2.34 for the two training budgets because fewer boundaries are stitched. In the right panel, switching the encoder or decoder to global processing gives only $\Delta\mathrm{CD}=+0.14/+0.19$, $\Delta\mathrm{ANC}=0.00/-0.03$, and $\Delta F=-0.07/-0.22$ relative to chunk-wise processing. Together, these small differences show that training and inference budgets, as well as encoder and decoder partitions, can be chosen independently without retraining.

\begin{table}[H]
\caption{Decoupling on HQM. Left: CD across training and inference budgets. Right: global encoder or decoder processing (dash) against the ${<}1$M encoder/${<}2$K decoder default.}
\label{tab:decoupling}
\centering
\small
\begin{minipage}{0.43\textwidth}
\centering
\setlength{\tabcolsep}{3pt}
\begin{tabular}{l cccc}
\toprule
\multirow{2}{*}{Train budget}&\multicolumn{4}{c}{Inference budget (CD)}\\
\cmidrule(lr){2-5}
&${<}1$M&${<}2$M&${<}4$M&${<}8$M\\
\midrule
${<}1$M&2.99&2.70&2.59&2.42\\
${<}100$K&3.04&2.71&2.57&2.34\\
\bottomrule
\end{tabular}
\end{minipage}\hfill
\begin{minipage}{0.53\textwidth}
\centering
\setlength{\tabcolsep}{3pt}
\begin{tabular}{ll ccc}
\toprule
Encoder budget&Decoder budget&CD&ANC&F\\
\midrule
${<}1$M&${<}2$K&2.99&99.04&98.89\\
--&${<}2$K&3.13&99.04&98.82\\
${<}1$M&--&3.18&99.01&98.67\\
\bottomrule
\end{tabular}
\end{minipage}
\end{table}

\paragraph{Stitched latent integrity.}
Figure~\ref{fig:features} provides a visual comparison: the global PCA structure remains stable across encoder budgets, while the difference insets localize residuals near chunk boundaries (highlighted by red circles), demonstrating the need for our proposed chunk stitching. Visual similarity alone, however, does not verify spatio-semantic integrity. We therefore adapt a PointNet classifier~\citep{qi2017pointnet} to accept 16-channel latent features and train it on global ModelNet40~\citep{wu20153d} latents extracted with unchunked VAE processing, achieving 99.52\% training accuracy and 76.61\% test accuracy. We then freeze the classifier and evaluate stitched latents without adaptation. Relative to no chunking, Table~\ref{tab:feature_integrity} reports $\Delta\mathrm{Accuracy}$ values of $-0.97$, $-0.36$, and $-0.24$ points for budgets below 1M, 2M, and 4M cells, respectively. Since the classifier never observes stitched latents during training, the nearly unchanged accuracy confirms that stitching preserves latent feature integrity for downstream tasks.

\FloatBarrier
\noindent\begin{minipage}[c]{0.40\textwidth}
\centering
\captionof{table}{Frozen ModelNet40 accuracy ($\times100$); $\Delta$ from no chunking.}
\label{tab:feature_integrity}
\small
\setlength{\tabcolsep}{3pt}
\raggedright
\begin{tabular}{lcc}
\toprule
Encoder processing&Accuracy&$\Delta$\\
\midrule
Chunk budget ${<}1$M&75.64&$-0.97$\\
Chunk budget ${<}2$M&76.25&$-0.36$\\
Chunk budget ${<}4$M&76.37&$-0.24$\\
No chunking&76.61&0.00\\
\bottomrule
\end{tabular}
\end{minipage}\hfill
\begin{minipage}[c]{0.56\textwidth}
\centering
\captionof{figure}{Latent PCA (purple) across encoder budgets; insets show differences to global encoding.}
\label{fig:features}
\includegraphics[width=\linewidth]{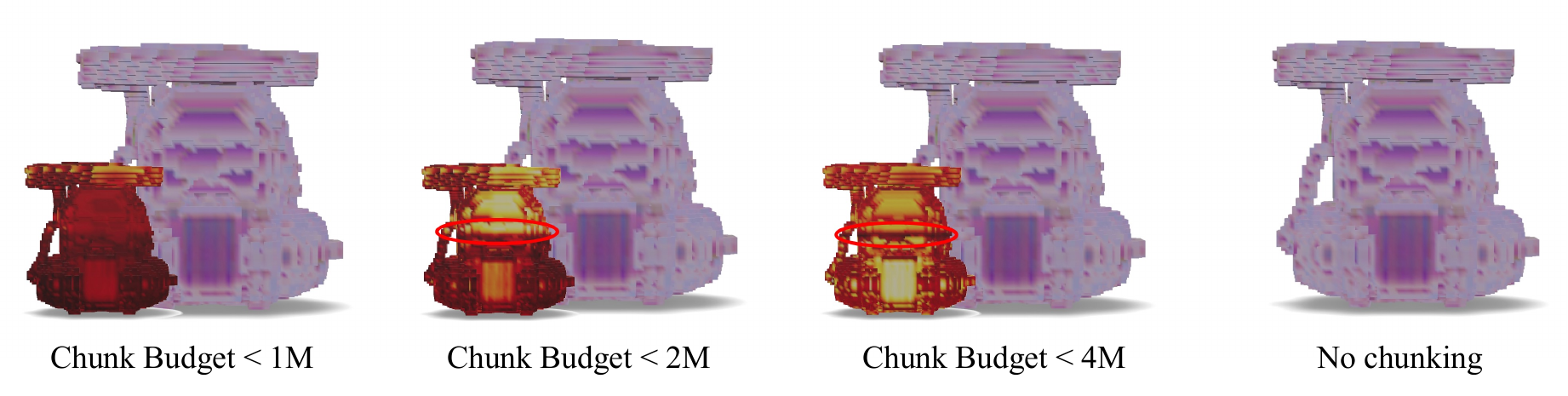}
\end{minipage}

\section{Scope, Limitations, and Conclusion}
\label{sec:conclusion}

\method{} makes sparse grid VAE computation local while preserving a stitched global latent. Local learned operators, \bbop{}, and \scurve{} stitching decouple training and inference budgets as well as encoder and decoder partitions. Experiments demonstrate state-of-the-art overall object reconstruction from $512^3$ to $1536^3$ and a strong balance between inference speed and adjustable memory. The ablations validate robust chunking, stitching, and latent integrity across partition choices. This preserved latent representation, in turn, supports state-of-the-art image to 3D generation.

Despite these results, the current formulation has two main limitations. \textit{(1) Watertight Constraint.} To reduce geometric artifacts in generated outputs, our current implementation uses watertight SDF geometry, which excludes open surfaces and some internal structures. Extending the local representation to non-watertight geometry while retaining reliable boundary assembly is an important direction, for example by applying \method{} to the o-voxel representation of TRELLIS.2~\citep{xiang2025native}. \textit{(2) Global DiT Training Bottleneck.} Although VAE encoding and decoding are chunk-wise, the downstream DiT still processes the stitched global latent, whose token count grows with global resolution. Fully scalable generation therefore requires extending the same locality principle to chunk-wise generative training and inference.

\label{page:mainend}
\clearpage
\phantomsection\label{page:refsstart}
\bibliographystyle{iclr2027_conference}
\bibliography{iclr2027_conference}

\clearpage
\appendix

\section{Implementation Details}
\label{app:implementation}

\method{} uses latent dimension 16. Encoder channels are [128, 256, 512, 512], and decoder channels are [512, 512, 256, 128]. The bottleneck Swin block has eight heads of dimension 64 and uses absolute positional embeddings. Each pruning head is a GELU-activated MLP~\citep{hendrycks2016gaussian}; teacher forcing supervises its loss using the corresponding intermediate encoder grid. Encoder and decoder padding are 32 and 4 cells, respectively.

Raw meshes are first repaired with heuristic hole filling and then converted into watertight surfaces before SDF computation. Flood filling identifies exterior empty cells, and least-squares optimization restores geometric details on the watertight surface. The surface field is then evaluated in a narrow band around the resulting boundary. Empirically, extending the band by two cells on each side prevents holes in reconstructed meshes without introducing excessive memory overhead.

\section{Partitioning and Stitching Algorithms}
\label{app:algorithms}

\begin{algorithm}[h]
\caption{Balanced Binary Object Partitioning}
\label{alg:bbop}
\begin{algorithmic}[1]
\Require sparse grid $X$, coordinates $C$, padding $P$, cell budget $B$
\State $B_{\mathrm{pad}}\gets\operatorname{Pad}(\operatorname{BBox}(C),P)$
\State $X_c\gets\{x\in X:x\in B_{\mathrm{pad}}\}$
\If{$|X_c|>B$}
    \State $a\gets\argmaxop_d(\max C_d-\min C_d)$
    \State $m\gets\operatorname{median}(C[:,a])$
    \State $C_L\gets\{c\in C:c_a\leq m\}$; $C_R\gets C\setminus C_L$
    \State \Return $\operatorname{BBOP}(X,C_L,P,B)\cup\operatorname{BBOP}(X,C_R,P,B)$
\Else
    \State \Return $\{X_c\}$
\EndIf
\end{algorithmic}
\end{algorithm}

\begin{algorithm}[h]
\caption{\scurve{} Weighted Stitching}
\label{alg:stitching}
\begin{algorithmic}[1]
\Require chunks $\{(C_i,F_i,B_i^{\mathrm{core}},B_i^{\mathrm{pad}})\}$
\State initialize global buffers $F_{\mathrm{sum}}$ and $W_{\mathrm{sum}}$
\For{each chunk $i$}
    \State retain coordinates $\hat C_i$ inside $B_i^{\mathrm{pad}}$
    \State compute $W_i=\prod_{d\in\{x,y,z\}}w(\hat C_{i,d})$ using Equation~\ref{eq:weight}
    \State scatter add $F_iW_i$ to $F_{\mathrm{sum}}$ and $W_i$ to $W_{\mathrm{sum}}$
\EndFor
\State \Return $F_{\mathrm{sum}}/W_{\mathrm{sum}}$ at occupied global coordinates
\end{algorithmic}
\end{algorithm}

\paragraph{Algorithmic behavior.}
BBOP separates recursive split coordinates from padded execution boxes. Median splits preserve coverage, padding supplies context, and padded stopping enforces $B$. Changing $B$ alters partition depth and boundary count, not the input grid or learned parameters. Stitching normalizes accumulated reliability to preserve scale and suppress padded-face predictions.

\section{Extended Qualitative Results}

\begin{figure}[p]
    \centering
    \includegraphics[width=\textwidth]{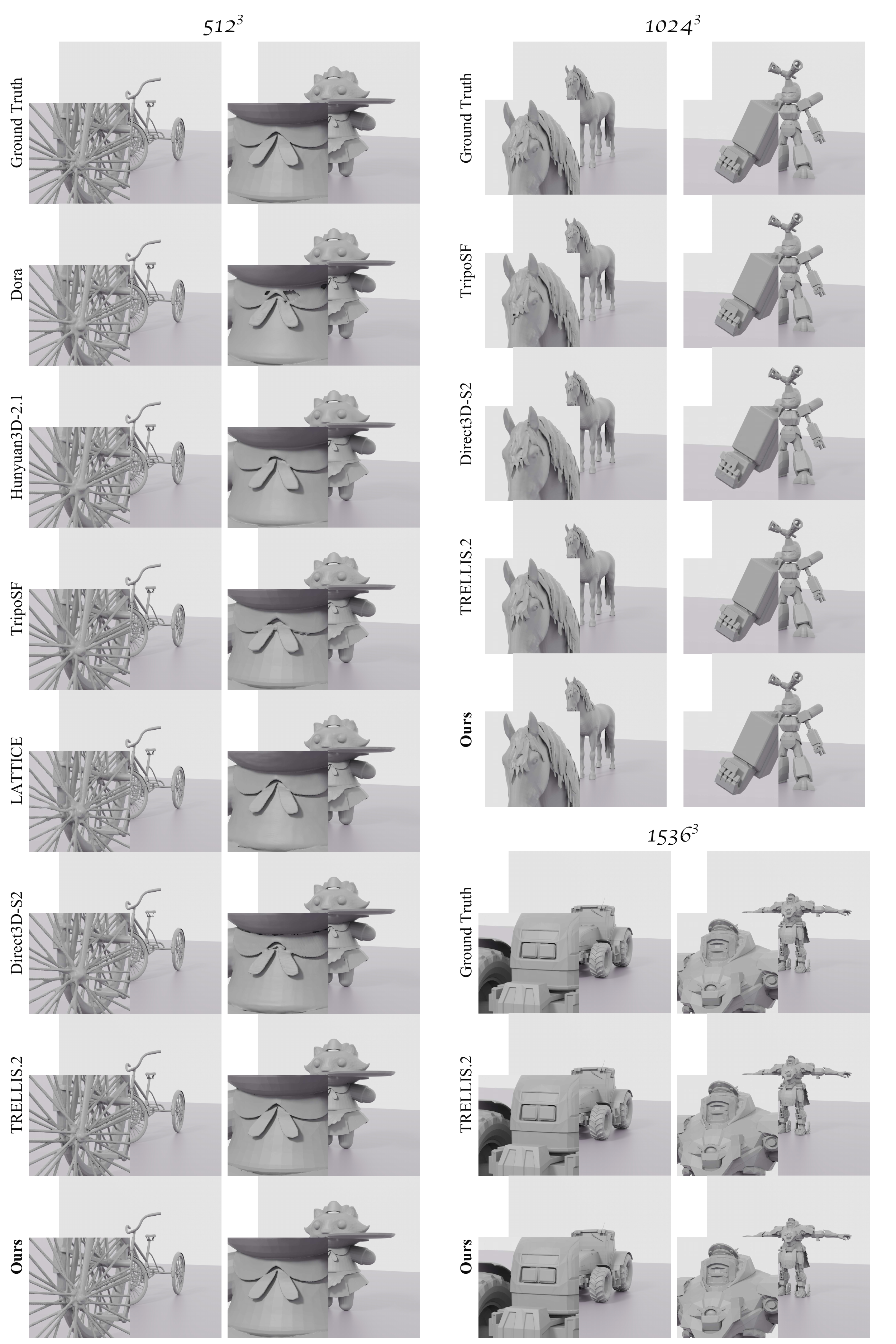}
    \caption{Additional qualitative VAE reconstruction on Toys4K at $512^3$, $1024^3$, and $1536^3$. \textit{Best viewed with zoom-in.}}
    \label{fig:more_reconstruction}
\end{figure}

\paragraph{Reconstruction.}
Figure~\ref{fig:more_reconstruction} localizes bicycle spokes and figurine ribbon detail at $512^3$, a horse mane and mechanical appendage at $1024^3$, and truck panels and mech parts at $1536^3$. Silhouettes remain stable; differences concentrate in thin, repeated, or high-curvature geometry. Our results remain close to the ground truth, consistent with the saturated F-scores.

\begin{figure}[p]
    \centering
    \includegraphics[height=0.88\textheight,keepaspectratio]{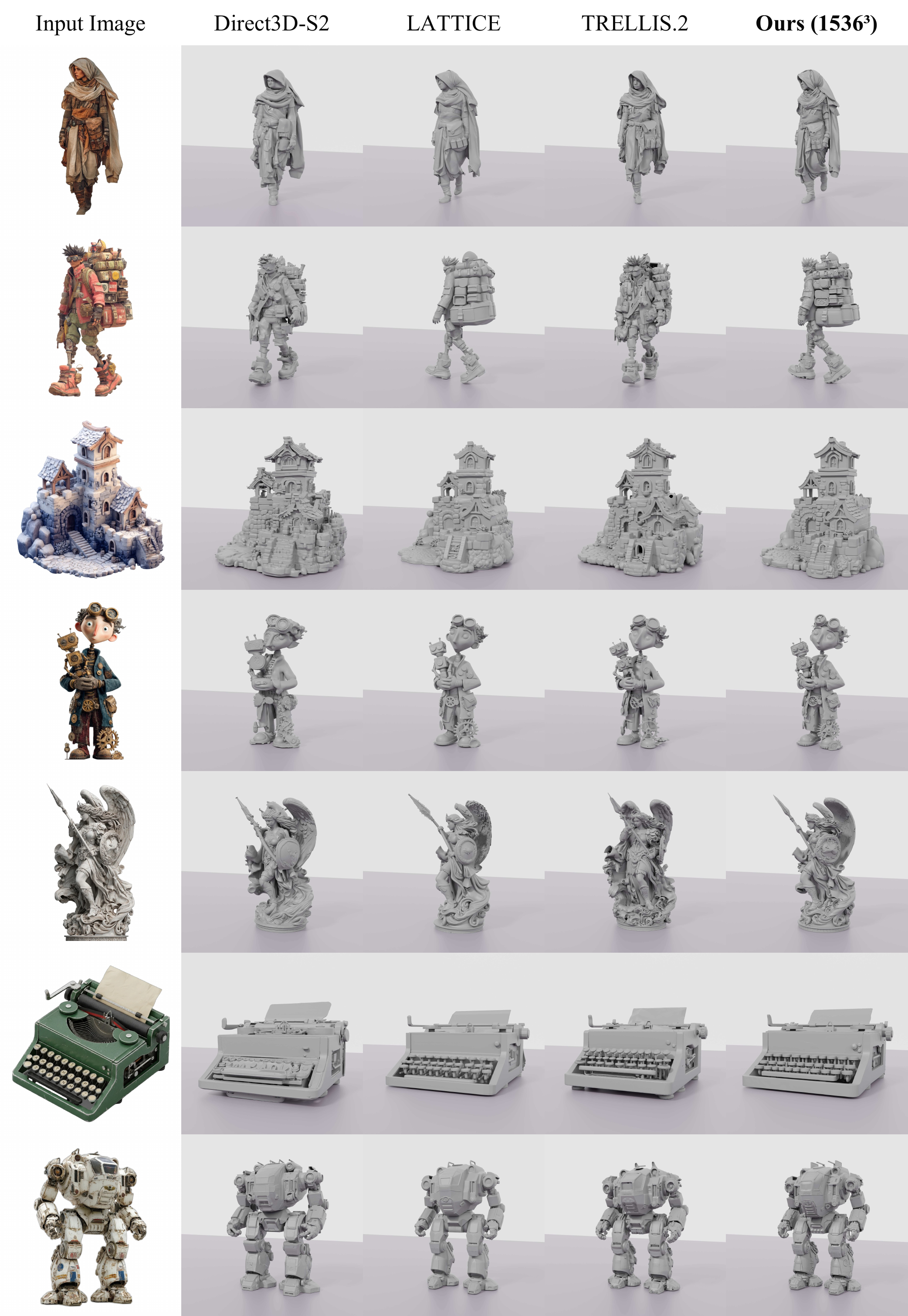}
    \caption{Additional image to 3D generation comparisons. \textit{Best viewed with zoom-in.}}
    \label{fig:more_generation}
\end{figure}

\noindent\textbf{Image-conditioned generation.}
Figure~\ref{fig:more_generation} spans diverse in-the-wild inputs. Compared with the baselines, our results align more closely with the input images, preserve finer details such as machine logos, sculptural details, and clothing accessories on cartoon characters, and produce more complete geometry without breakage. In contrast, TRELLIS.2 exhibits frequent fractures and holes.

\clearpage
\clearpage

\clearpage

\end{document}

%% file: math_commands.tex
\usepackage{amsmath,amsfonts,bm}

\def\eqref#1{equation~\ref{#1}}
\def\1{\bm{1}}

\DeclareMathAlphabet{\mathsfit}{\encodingdefault}{\sfdefault}{m}{sl}
\SetMathAlphabet{\mathsfit}{bold}{\encodingdefault}{\sfdefault}{bx}{n}